\documentclass{article}

\usepackage[dblblindworkshop, final]{neurips_2026}
\usepackage[utf8]{inputenc} 
\usepackage[T1]{fontenc}    
\usepackage{hyperref}       
\usepackage{url}            
\usepackage{booktabs}       
\usepackage{amsfonts}       
\usepackage{nicefrac}       
\usepackage{microtype}      
\usepackage{xcolor}         

\usepackage{tikz}
\usetikzlibrary{arrows.meta}
\usetikzlibrary{patterns}
\usepackage{amsmath}
\usepackage{multirow}
\usepackage{graphicx}

\title{Primitives and Paths: Predicting What Post-Training Can Do from What the Base Model Already Has}
\workshoptitle{Transitioning from Pre-Training to Post-Training}

\author{%
  Anqi Qu \\
  University of Oxford \\
  \texttt{anqi.qu@ccc.ox.ac.uk} 
}

\begin{document}

\maketitle

\begin{abstract}
Does reinforcement learning teach language models new capabilities, or merely amplify ones acquired during pre-training? The literature offers evidence for both: pass@$k$ studies suggest RL reweights latent behaviours while some controlled composition studies demonstrate gains beyond simple amplification. We argue that this question has no answer independent of the capability state of the base model, and propose a two-dimensional state space for post-trainability. The first axis, \emph{effective full-solution support}, measures whether successful trajectories can be sampled under a feasible exploration budget. The second axis, \emph{atomic primitive coverage}, measures whether the constituent skills of a solution can be executed when the burden of composition is removed. Together, these dimensions motivate three regimes -- amplification, composition-limited, and acquisition-limited -- each predicting a distinct empirical signature and calling for a distinct intervention. We propose a prospective measurement protocol and investigate the framework using synthetic composition tasks across three small language models. We find that primitive mastery alone does not guarantee that RL can learn a composition. Checkpoints that execute every primitive reliably can fail to improve when complete solutions are never sampled during training, whereas modest composition supervision that establishes nonzero support enables substantial RL gains. The framework reframes mid-training as preparation for post-training: its role is to move models into states from which RL can operate effectively.
\end{abstract}

\section{Introduction}

Reinforcement learning with verifiable rewards (RLVR) has become a central component of post-training language models for mathematical, coding, and other reasoning-intensive tasks. Yet the reasons behind its empirical success remain poorly understood and there is substantial disagreement over what RL contributes beyond capabilities already acquired during pre-training. Does RL teach models genuinely new ways to reason, or does it primarily make existing but low-probability behaviours more likely?

Recent evidence supports both views. \cite{yue_does_2025} show that although RL-trained models substantially outperform their base models at pass@1, this advantage often disappears at sufficiently large sampling budgets: given enough attempts, the base model recovers many of the same successful reasoning trajectories as its RL-trained counterpart. \cite{zhao_echo_2025} similarly find that RL strongly amplifies behaviours established by the pre-training distribution. Together, these results suggest a view of RL as primarily reweighting latent capabilities rather than expanding them.

Other results complicate this position. \cite{yuan_fx_2025} construct tasks in which constituent skills are available to the base model while their composition is withheld, and show that RL can learn new, transferable compositions of those skills. \cite{chu_sft_2025} find that RL can generalise beyond training examples in settings where SFT is prone to memorisation, while \cite{bansal_excursions_2026} find that direct RL can expand high-$k$ performance when applied to pre-training checkpoints, even in settings where the conventional SFT-then-RL pipeline instead sharpens the output distribution. Taken together, these findings suggest that ``does RL create new capabilities?'' may not have a single answer independent of the model's state before post-training.

We propose that the effect of post-training should instead be understood relative to the capability state of the base model. In particular, we distinguish two properties of a base model on a task: 
\begin{enumerate}
    \item \textbf{Full-solution support}: the extent to which complete successful trajectories are already represented in the model's output distribution
    \item \textbf{Atomic primitive coverage}: the extent to which the constituent skills required for a successful solution are individually available, even when the model cannot reliably compose them
\end{enumerate}
These dimensions distinguish qualitatively different post-training regimes. When both are high, RL can amplify already successful behaviour. When atomic primitive coverage is high but complete solutions are rare, post-training faces a composition problem. When the required atomic primitives themselves are absent, the task is instead acquisition-limited and may require additional supervised or mid-training data before RL can be effective.

This perspective is motivated by a growing body of work showing that standard pre-training metrics are insufficient for predicting downstream post-training outcomes. \cite{kang_quagmires_2025}, for example, find that high pre-RL SFT performance can be negatively associated with final RL performance and that pass@large-$k$ and held-out generalisation loss provide substantially stronger predictors. Theoretically, \cite{chen_coverage_2025} formalise a related notion of coverage: the probability mass a pre-trained model assigns to high-quality responses, and show that coverage plays a fundamental role in determining the effectiveness of subsequent post-training and inference-time search. At the same time, decomposing overall task accuracy into constituent skills can reveal important capability gaps obscured by aggregate metrics (\cite{qin_decomposing_2025}).

Our goal is therefore to develop a diagnostic framework for post-trainability. Given a base model and a target problem distribution, what can we measure before post-training that tells us which learning regime we are in and how effective subsequent RL is likely to be? We operationalise full-solution support and atomic primitive coverage, study whether these measurements predict post-training outcomes, and use the resulting state space to propose implications for mid-training. More broadly, we argue that mid-training should not be viewed solely as a means of maximising base-model performance. Its role may instead be to move models into capability states from which subsequent post-training can operate effectively.
\section{Background}
\subsection{RL as capability amplification}
One prominent view of reasoning post-training is that RL primarily increases the probability of successful behaviours that are already latent in the base model. \cite{yue_does_2025} motivate this view through pass@$k$. While post-RL models outperform their corresponding base models when only a small number of samples are drawn, base models often catch up to the post-RL models at large $k$. Under this interpretation, pre-training has already placed some probability mass on successful reasoning trajectories. RL changes their relative likelihood and converts rare successes into reliable behaviour.

\cite{zhao_echo_2025} provide complementary evidence by controlling pre-training data mixtures. Across different RL algorithms and model scales, they find that RL tends to concentrate generation around behavioural patterns already induced by pre-training. Changes to the pre-training distribution consequently alter the behaviours that post-training ultimately amplifies. This establishes an important dependence between the two stages: the same post-training objective can produce different outcomes depending on what pre-training has made available.

\cite{chen_coverage_2025} define the \emph{coverage} of a pre-trained model in terms of the probability mass assigned to high-quality responses. They show that coverage, rather than cross-entropy loss alone, influences the ability of several downstream procedures to recover good outputs. \cite{foster_foundation_2025} similarly establish a computational role for base-model coverage in efficient exploration. These results motivate measuring not only what a base model produces on its first attempt, but also the structure of the wider distribution from which post-training begins.

Large-$k$ pass rates are a natural empirical proxy for this property but should not be interpreted as a perfect measure of reasoning support. In particular, large sampling budgets can eventually produce accidental correct answers on tasks with small or discrete answer spaces. We therefore treat pass@$k$ as one operational measure of full-solution support rather than identifying it with support itself.

\subsection{Atomic skills, composition, and skill acquisition}
A pure reweighting account of RL cannot explain all observed RL gains. \cite{yuan_fx_2025} explicitly separate possession of atomic skills from the ability to compose them. Their models are first trained on individual functions such as $f$ and $g$ while the target composition $f(g(x))$ is withheld. Subsequent RL can help the model acquire the unseen composition and transfer the learned compositional behaviour to new settings. This suggests that the absence of complete successful trajectories does not imply that a task lies outside the model's post-trainable capability boundary because the relevant ingredients may already exist in decomposed form.

\cite{qin_decomposing_2025} similarly decompose mathematical problem solving into planning, execution, and verification capabilities, and find that RL can affect these components differently. In their experiments, RL predominantly improves execution robustness on problems the model already knows how to solve, while weaknesses in planning impose a ``coverage wall'' on more novel problems. Whether post-training succeeds may thus depend not only on whether a complete solution is ever sampled, but on which constituent skills are available and which bottleneck prevents them from being assembled into a successful trajectory.

Pre-existing behaviours also shape a model's capacity to self-improve. \cite{gandhi_cognitive_2025} identify verification, backtracking, subgoal setting, and backward chaining as behaviours associated with successful self-improvement under RL. Models initially lacking these behaviours improve less under identical post-training, whereas exposure to data rich in these reasoning patterns -- including through continued pre-training -- substantially enhances subsequent gains. Two pretrained models with similar headline accuracy may therefore present very different substrates for post-training.

\subsection{The interaction between SFT, RL, and model state}
The appropriate post-training objective may itself depend on the capability state of the base model. \cite{chu_sft_2025} find that SFT and RL play different roles: SFT is useful for establishing output conventions and initial behaviours while outcome-based RL generalises more strongly in controlled tasks. Conversely, when the model cannot obtain useful rewards on difficult examples, SFT can provide capabilities that RL struggles to discover. This motivates methods such as ReLIFT \cite{ma_relift_2026}, which selectively introduces supervised solutions for questions outside the model's current capability range.

Difficulty therefore matters relative to the current policy rather than solely as an intrinsic property of a problem. \cite{hu_dereason_2026} study how examples of varying difficulty should be allocated across SFT and RL, showing that the allocation materially affects final reasoning performance. \cite{kang_quagmires_2025} further show that maximising SFT performance is not equivalent to finding the best initialisation point for RL: checkpoints with stronger SFT scores (higher accuracy) can produce weaker final models after identical RL training. Together, these findings motivate evaluating a checkpoint by its \emph{post-training potential}, rather than its immediate accuracy alone.

\cite{bansal_excursions_2026} push this observation further by applying RL directly to intermediate pre-training checkpoints. They find that RL can improve reasoning substantially even relatively early in pre-training, and that the qualitative effect of RL depends on the preceding training pipeline. In their experiments, SFT followed by RL tends to sharpen the output distribution whereas direct RL can improve both pass@1 and higher-$k$ performance. ``What RL does'' thus cannot be separated cleanly from the state from which it begins.

\subsection{Mid-training as preparation for post-training}
If successful post-training depends on particular skills or behaviours already being available, then mid-training offers an opportunity to deliberately establish the prerequisites that RL can subsequently exploit. Recent results support this interpretation. \cite{zhang_interplay_2025} jointly study pre-training, mid-training, and RL in controlled reasoning environments. They find that RL gains depend strongly on whether target tasks lie near the model's existing ``competence boundary'' and that allocating part of a fixed compute budget to mid-training can substantially improve subsequent RL generalisation. \cite{gandhi_cognitive_2025} similarly show that continued pre-training on data enriched for particular reasoning behaviours can transform a model that plateaus under RL into one that self-improves effectively.

This suggests a different objective for mid-training. Rather than asking only which additional data maximise pre-RL benchmark accuracy, we may ask which data establish the \emph{primitives and behavioural support} required for successful post-training later. Under this view, reasoning-rich mid-training and RL have complementary roles: mid-training can supply or strengthen missing atomic skills, while RL can exploit, compose, and amplify those skills through on-policy experience. 

We introduce a state-space framework in the following section that makes this distinction explicit and provides an operational basis for deciding which intervention is appropriate for a given model-task pair.

\definecolor{qComp}{RGB}{224,234,247}   
\definecolor{qAmp}{RGB}{225,240,227}    
\definecolor{qAcq}{RGB}{244,229,229}    
\definecolor{qDiag}{RGB}{239,239,239}   

\section{The State Space}
\paragraph{Full-solution support.} The first dimension captures the extent to which a base model can already generate complete successful solutions before post-training. This is closely related to the notion of coverage studied by \cite{chen_coverage_2025}, which measures the probability mass a pretrained model places on correct or high-quality responses. Large-$k$ sampling provides one way to empirically probe this property because a problem with low pass@1 but high pass@$k$ reveals successful trajectories that the base policy rarely selects, whereas a problem that remains unsolved at large $k$ provides little evidence that a complete solution is accessible to the model at all \cite{yue_does_2025}. 

We use ``support'' here in an operational sense rather than a strictly measure-theoretic sense. Finite sampling cannot distinguish a trajectory with zero probability from one with very small probability. Our quantity of interest is therefore \emph{effective} full-solution support: whether successful trajectories occur with sufficient probability to be recoverable under a feasible sampling and exploration budget. Crucially, this quantity depends on the decoding policy and computational budget, so failure to observe a successful trajectory does not imply that its probability is zero.

\paragraph{Primitive coverage.} The second dimension asks whether the model possesses the constituent skills required by the task when the burden of composing those skills is removed. Consider a solution decomposed into primitives $\mathcal{P}(x) = (s_1,\ldots,s_m)$.  A model may fail to generate the complete solution while nevertheless executing each $s_j$ reliably in isolation or when given the appropriate intermediate state or scaffold. We refer to the extent to which these constituent skills are available as \emph{atomic primitive coverage}. The distinction is motivated by evidence that aggregate accuracy can obscure substantial differences between capabilities such as planning, execution, and verification \cite{qin_decomposing_2025}, and that RL can learn unseen compositions of atomic skills that were already acquired before RL began \cite{yuan_fx_2025}. 

\paragraph{Relationship between the axes.}
Although the two dimensions capture distinct properties, they are not necessarily independent. A complete successful solution must, by construction, involve some valid sequence of constituent operations. We would therefore generally expect high full-solution support to be accompanied by the availability of the relevant primitives, while high primitive coverage need not imply high full-solution support since the model may fail to select or compose those primitives autonomously.

This suggests an asymmetry in the state space, with model-task pairs plausibly concentrated away from the lower-right region of the plane. We therefore interpret the state space as a diagnostic framework for identifying potential bottlenecks, rather than as four independent or sharply bounded regions.

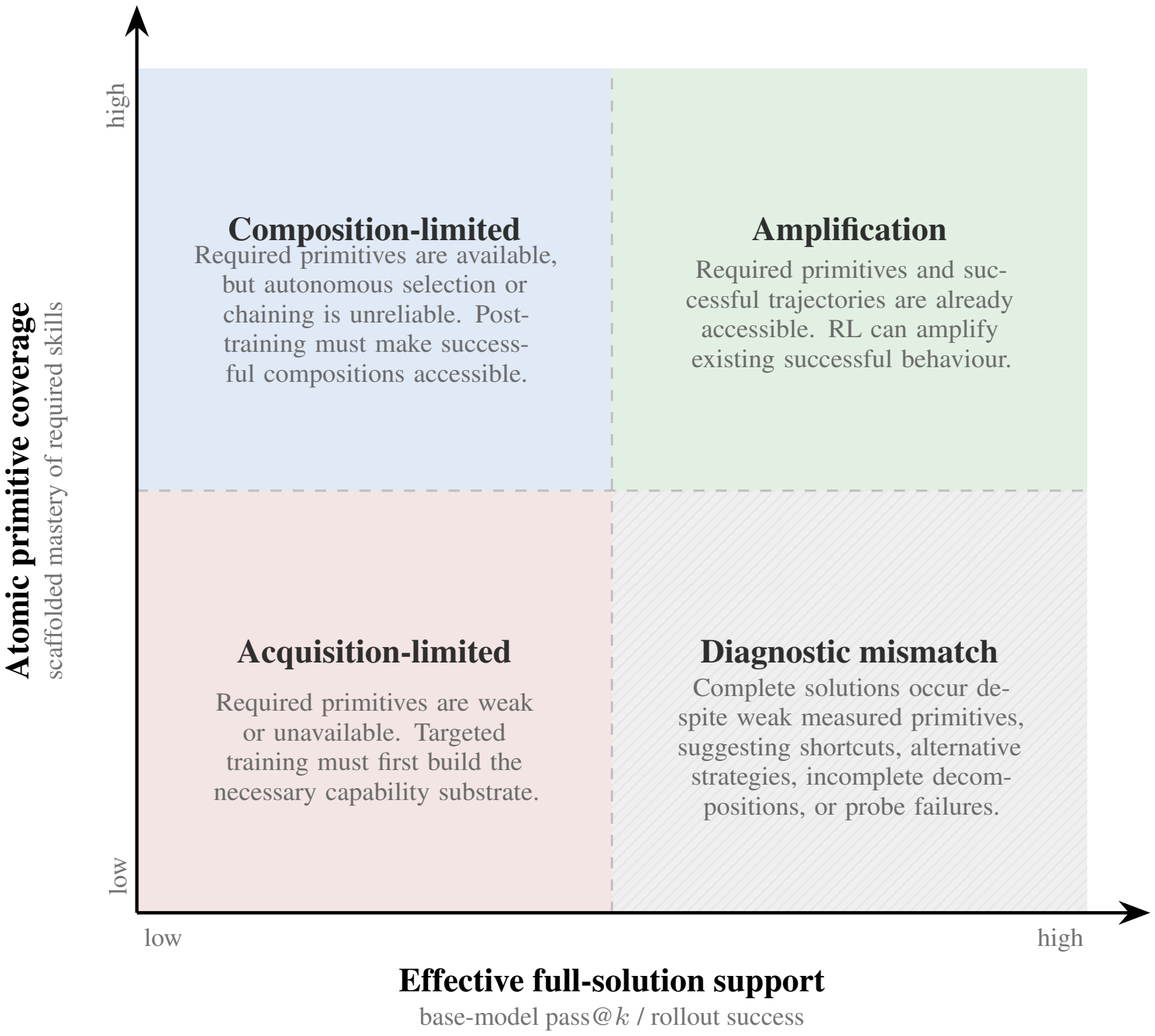
\begin{figure}[h]
    \centering

    \begin{tikzpicture}[
      font=\small,
      qtitle/.style={
        font=\small\bfseries,
        text=black!82,
        align=center
      },
      qdesc/.style={
        font=\scriptsize,
        text=black!58,
        align=center,
        text width=3.5cm
      },
      axislabel/.style={
        font=\small,
        align=center
      }
    ]
    
    \fill[qComp] (0,4) rectangle (4.5,8);
    \fill[qAmp]  (4.5,4) rectangle (9,8);
    \fill[qAcq]  (0,0) rectangle (4.5,4);
    \fill[qDiag] (4.5,0) rectangle (9,4);
    
    \fill[
      pattern=north east lines,
      pattern color=black!10
    ] (4.5,0) rectangle (9,4);
    
    \draw[black!25, dashed, line width=0.6pt] (4.5,0) -- (4.5,8);
    \draw[black!25, dashed, line width=0.6pt] (0,4) -- (9,4);
    
    \draw[-{Stealth[length=3mm]}, line width=0.9pt]
      (0,0) -- (9.6,0);
    
    \draw[-{Stealth[length=3mm]}, line width=0.9pt]
      (0,0) -- (0,8.6);
    
    \node[axislabel] at (4.5,-0.82) {
      \textbf{Effective full-solution support}\\[-1pt]
      {\scriptsize\color{black!55} base-model pass@$k$ / rollout success}
    };
    
    \node[axislabel, rotate=90] at (-0.95,4) {
      \textbf{Atomic primitive coverage}\\[-1pt]
      {\scriptsize\color{black!55} scaffolded mastery of required skills}
    };
    
    \node[below, font=\scriptsize, text=black!55]
      at (0.25,-0.02) {low};
    
    \node[below, font=\scriptsize, text=black!55]
      at (8.75,-0.02) {high};
    
    \node[font=\scriptsize, text=black!55, rotate=90]
      at (-0.18,0.35) {low};
    
    \node[font=\scriptsize, text=black!55, rotate=90]
      at (-0.18,7.65) {high};
    
    \node[qtitle] at (2.25,6.45)
      {Composition-limited};
    
    \node[qdesc] at (2.25,5.65)
      {Required primitives are available, but autonomous selection or chaining is unreliable. Post-training must make successful compositions accessible.};
    
    \node[qtitle] at (6.75,6.45)
      {Amplification};
    
    \node[qdesc] at (6.75,5.65)
      {Required primitives and successful trajectories are already accessible. RL can amplify existing successful behaviour.};
    
    \node[qtitle] at (2.25,2.45)
      {Acquisition-limited};
    
    \node[qdesc] at (2.25,1.55)
      {Required primitives are weak or unavailable. Targeted training must first build the necessary capability substrate.};
    
    \node[qtitle] at (6.75,2.45)
      {Diagnostic mismatch};
    
    \node[qdesc] at (6.75,1.55)
      {Complete solutions occur despite weak measured primitives, suggesting shortcuts, alternative strategies, incomplete decompositions, or probe failures.};

    \end{tikzpicture}
    \caption{A state space for post-trainability. Full-solution support
    measures whether complete successful trajectories are accessible under
    the base policy, while atomic primitive coverage measures whether the
    constituent skills required for a solution are individually available.}
    \label{fig:posttraining-state-space}
\end{figure}

Together, these two dimensions produce the state space shown in Figure~\ref{fig:posttraining-state-space}.

\subsection{Post-training regimes} 

\paragraph{Amplification.} When both primitive coverage and full-solution support are high, successful reasoning trajectories are already accessible under the base policy and rest on capabilities the model can reliably execute. Post-training does not need to construct a qualitatively new solution procedure and can instead increase the probability assigned to successful trajectories. This regime corresponds most directly to the reweighting account of RLVR proposed by \cite{yue_does_2025} and \cite{zhao_echo_2025}. This region can be thought of as ``easy RL'' although the difficulty of RL within this region still depends on the density of successful trajectories and different sampling budgets: a solution sampled once in $10^5$ attempts and one sampled once in five attempts are both cases of amplification but present very different exploration problems. When successful trajectories are sufficiently accessible, standard outcome-based RL can amplify them, typically producing larger improvements at small sampling budgets than at large ones. A characteristic empirical signature of amplification is therefore a substantial increase in low-$k$ performance after RL with comparatively little expansion at large $k$.

\paragraph{Composition-limited.} When primitive coverage is high but full-solution support is low, the model appears to possess the ingredients of a solution without reliably being able to assemble them into an end-to-end trajectory. We call this regime \emph{composition-limited}. \cite{yuan_fx_2025} provide a controlled example of this setting: models that had acquired atomic functions $f$ and $g$ could subsequently learn the withheld composition $f(g(x))$ through RL. 

However, possessing the required primitives does not guarantee that vanilla outcome-based RL can learn their composition. If successful end-to-end trajectories are too rare under the training-time sampling distribution, the model may receive little or no informative reward signal even when it can execute every primitive individually. Successful post-training would constitute more than simple probability sharpening here. If complete solutions that were absent or extremely rare before RL become recoverable at large sampling budgets afterwards, while the underlying primitives were already mastered, this is evidence of compositional learning. When compositional learning succeeds, we expect improvements in both pass@1 and effective full-solution support. Interventions in this regime should therefore target the accessibility and execution of complete compositions, potentially through outcome-based RL when rewards are reachable, or through composition-focused supervision, curricula, scaffolding, or intermediate rewards when they are not.

\paragraph{Acquisition-limited.} When both full-solution support and primitive coverage are low, at least part of the substrate required for the target task appears to be missing or unreliable. We refer to this regime as \emph{acquisition-limited}. Low measured primitive coverage does not necessarily establish that a skill is absent. It may instead reflect an inappropriate decomposition, insufficient scaffolding, or a mismatch between the probing context and the context in which the skill is actually available. Vanilla outcome-based RL faces two difficulties here: successful trajectories provide little reward signal, and effective exploration may still fail if the model cannot execute one or more necessary constituent skills. This regime motivates interventions before or alongside RL. Mid-training, SFT, or distillation can strengthen missing or unreliable primitives, potentially increasing primitive coverage and making subsequent composition learning more feasible.

\paragraph{Diagnostic mismatch.} High measured full-solution support combined with low measured primitive coverage is qualitatively different from the other regions. If a model reliably solves a task while failing probes for the skills assumed to be necessary, the most immediate conclusion would be to call the decomposition or measurement procedure into question. The model may be exploiting an alternative solution strategy, relying on memorised or contaminated examples, using a shortcut not captured by the proposed primitives, or exposing weaknesses in the primitive probes themselves. We therefore interpret this region as a prompt for further diagnosis rather than a distinct learning regime or a mathematically impossible state.

\paragraph{A continuous state space.} The three learning regimes and one diagnostic region are intended as conceptual regimes rather than hard categories. Both axes are continuous, and the relevant transition points depend on the post-training algorithm, rollout budget, verifier, and task. Our central claim is therefore not that every model-task pair belongs unambiguously to one quadrant, but that these two measurements expose different bottlenecks that aggregate base-model accuracy conflates.

\section{Measurement Protocol}
We propose a protocol for locating a model-task pair in the state space before post-training. The protocol consists of obtaining independent reference decompositions, measuring effective full-solution support and atomic primitive coverage, diagnosing compositional bottlenecks, and assigning provisional regimes. All measurements and classifications should be made before observing post-training outcomes. Throughout, let $\pi_0$ denote the base policy and $\pi_1$ a post-trained policy derived from it.

The central obstacle in measuring primitive coverage is obtaining a valid decomposition $\mathcal{P}(x)$ without the synthetic control of \cite{yuan_fx_2025}. Some ways that reference solutions can be obtained independently of $\pi_1$ are using human annotations, stronger external models, or programmatically generated solutions where available. Each reference trajectory is verified for correctness and decomposed into an ordered sequence of primitives $\mathcal{P}(x) = (s_1,\ldots,s_m)$. Where multiple valid decompositions exist, these should all be considered when assessing the robustness of primitive coverage.

\paragraph{Estimating full-solution support.}
For each problem $x$, we draw $n$ independent samples from $\pi_0$ under a specified decoding policy and estimate pass@$k$ for $k \le n$ using the unbiased estimator of \cite{chen_evaluating_2021}. We report both the per-rollout success rate, which estimates the probability of a successful individual trajectory, and pass@$k$, which measures success under a sampling budget of $k$ attempts. The problem's $x$-coordinate is $\widehat{p}_0(x) = \mathrm{pass@}K(x)$ at the largest feasible budget $K$. Since final-answer matching can overestimate support on tasks where answers can be guessed, we follow \cite{yue_does_2025} and validate a random subsample of successful base-model trajectories for reasoning validity, reporting the estimated false-success rate.

\paragraph{Estimating primitive coverage.}
For each primitive $s_j$ in the reference decomposition, we can construct an isolated execution probe which supplies the appropriate intermediate state and explicitly specifies the operation to be performed, removing the burden of selecting the next primitive. Let $q_j^{\mathrm{exec}}$ denote the probability that $\pi_0$ executes this operation correctly.

We define atomic primitive coverage using the isolated execution probabilities:
$$
\widehat{c}(x) = \min_j q_j^{\mathrm{exec}}(x),
$$
since a single unreliable required primitive can create an execution bottleneck. We additionally report aggregate primitive success rates as the minimum is sensitive to probe noise and decomposition granularity.

\paragraph{Scaffolding and composition diagnostics.}
To investigate this distinction, we compare autonomous solution success with completion under oracle assistance. A complementary instrument is the \emph{handoff curve}: given a verified reference trajectory, we supply the first $t$ steps to $\pi_0$ and measure its probability of correctly completing the remaining trajectory.

Let $H_x(t)$ denote this oracle-assisted completion probability, with $H_x(0)$ corresponding to autonomous full-solution success. We define an empirical composition gap at handoff point $t$ as
$$
G_{\mathrm{comp}}(x;t) = H_x(t) - H_x(0).
$$

A large positive gap indicates that supplying part of the solution substantially improves completion which is consistent with a bottleneck in autonomous composition. Comparing different scaffolds, such as supplying the next operation explicitly or providing only the current intermediate state, can further help distinguish operation selection from primitive execution and multi-step chaining.

\paragraph{Classifying model-task pairs.}
We assign provisional regimes using the pre-training-time coordinates $(\widehat{p}_0(x),\widehat{c}(x))$. A pair is classified as amplification-oriented when both coordinates exceed thresholds $\tau_p$ and $\tau_c$, composition-limited when support is below $\tau_p$ but primitive coverage exceeds $\tau_c$, and potentially acquisition-limited when both are below threshold.

We should treat $\tau_p$ and $\tau_c$ as reporting parameters rather than universal constants. They should be specified before post-training and varied in sensitivity analyses. In particular, the support threshold should be interpreted relative to the rollout budget and sampling distribution of the post-training algorithm. 

\paragraph{Triage for measurement mismatch.}
High measured support with low measured coverage is handled by inspecting the successful trajectories themselves, in decreasing order of parsimony. If they follow a decomposition different from $\mathcal{P}(x)$, the primitive set was wrong and should be re-probed against the observed decomposition. If they follow $\mathcal{P}(x)$ while the corresponding scaffolded probes fail, the probes are at fault and should be repaired. If the trajectories are themselves invalid (e.g.\ answer-first rationalisation, contamination artifacts, or reasoning that does not support the answer), the observation lends itself to a shortcut interpretation, predicting poor transfer. Each exit either relocates the problem or repairs the instrument; none defines a training regime.
\section{Testing the State-Space Framework}

We test our proposed framework by constructing a synthetic composition task in which supervised training controls exposure to individual primitives and their compositions. We then measure each checkpoint's pre-RL primitive coverage and effective full-solution support and test whether these measurements predict its response to identical outcome-based RL. To distinguish effects of training history from model architecture, we repeat the experiment across three small language models: Qwen3-1.7B, Gemma-3-1B, and Qwen2.5-1.5B-Instruct.

\paragraph{Task and setup.}
Following the setting of \cite{yuan_fx_2025}, we define two operations on five-digit strings: $F$ rotates the string left by one position, and $G$ increments its first digit by three modulo ten. Each problem requires a sequence of 2-4 operations containing both $F$ and $G$. A programmatic verifier checks every intermediate state and the final answer.

For each architecture, we construct checkpoints with different
supervised training histories:
\begin{enumerate}
    \item $\bold{M_1}$ is trained on $F$ only.
    \item $\bold{M_1{+}C^{(n)}}$ starts from $M_1$ and receives $n$ examples of mixed compositions, so it sees $G$ only inside compositions.
    \item $\bold{M_2}$ is trained on both $F$ and $G$, but never on mixed compositions.
    \item $\bold{M_2{+}C^{(n)}}$ starts from $M_2$ and receives the same $n$ composition examples as $M_1{+}C^{(n)}$.
\end{enumerate}
In both composition conditions, $n \in \{100,300\}$ and the models are trained for one epoch.

These checkpoints are designed to represent different regions of our state space. $M_1$ represents an acquisition-limited state since one of the required primitives is unavailable. The comparison between $M_1$ and $M_2$ represents a primitive-acquisition intervention. In $M_2$, both primitives are now available, but their autonomous composition remains unreliable, which places $M_2$ in the composition-limited regime. Adding composition supervision ($M_2 \rightarrow M_2{+}C^{(n)}$) makes complete trajectories more accessible to RL. The model thus moves towards the amplification regime in the $M_2{+}C^{(n)}$ checkpoints.

We measure pre-RL success on 150 held-out problems using 64 samples per problem. We then apply the same outcome-based GRPO procedure to each checkpoint (150 updates, three seeds). Rewards require a fully correct trajectory.

\begin{table*}[h]
\centering
\small
\renewcommand{\arraystretch}{1.12}
\begin{tabular*}{\textwidth}{@{\extracolsep{\fill}}clrrcrrrr@{}}
\hline
Model & Checkpoint &
\multicolumn{3}{c}{Pre-RL} &
\multicolumn{2}{c}{Post-RL} &
\multicolumn{2}{c}{Change} \\
 & & pass@1 & pass@64 & $F$ / $G$. & pass@1 & pass@64
 & $\Delta$pass@1 & $\Delta$pass@64 \\
\hline
\multirow{6}{*}{\rotatebox[origin=c]{90}{Qwen3-1.7B}}
 & $M_1$             & 0.000 & 0.000 & 1.00 / 0.00 & 0.000 & 0.000 & 0.000 & 0.000 \\
 & $M_1{+}C^{(100)}$ & 0.000 & 0.000 & 1.00 / 0.00 & 0.000 & 0.000 & 0.000 & 0.000 \\
 & $M_1{+}C^{(300)}$ & 0.003 & 0.033 & 0.97 / 0.00 & 0.172 & 0.376 & $+0.169$ & $+0.342$ \\
 & $M_2$             & 0.000 & 0.000 & 1.00 / 1.00 & 0.000 & 0.000 & 0.000 & 0.000 \\
 & $M_2{+}C^{(100)}$     & 0.064 & 0.180 & 1.00 / 1.00 & 0.452 & 0.482 & $+0.388$ & $+0.302$ \\
 & $M_2{+}C^{(300)}$     & 0.751 & 0.893 & 1.00 / 1.00 & 0.988 & 1.000 & $+0.237$ & $+0.107$ \\
\hline
\multirow{6}{*}{\rotatebox[origin=c]{90}{Gemma-3-1B}}
 & $M_1$             & 0.000 & 0.000 & 1.00 / 0.00 & 0.000 & 0.000 & 0.000 & 0.000 \\
 & $M_1{+}C^{(100)}$ & 0.000 & 0.000 & 0.99 / 0.00 & 0.000 & 0.000 & 0.000 & 0.000 \\
 & $M_1{+}C^{(300)}$ & 0.001 & 0.020 & 1.00 / 0.00 & 0.234 & 0.242 & $+0.232$ & $+0.222$ \\
 & $M_2$             & 0.000 & 0.000 & 1.00 / 1.00 & 0.476 & 0.500 & $+0.476$ & $+0.500$ \\
 & $M_2{+}C^{(100)}$     & 0.548 & 0.633 & 1.00 / 1.00 & 0.979 & 0.991 & $+0.431$ & $+0.358$ \\
 & $M_2{+}C^{(300)}$     & 0.960 & 0.987 & 1.00 / 1.00 & 0.989 & 0.998 & $+0.030$ & $+0.011$ \\
\hline
\multirow{6}{*}{\rotatebox[origin=c]{90}{Qwen2.5-1.5B}}
 & $M_1$             & 0.000 & 0.000 & 1.00 / 0.00 & 0.000 & 0.000 & 0.000 & 0.000 \\
 & $M_1{+}C^{(100)}$ & 0.000 & 0.000 & 1.00 / 0.00 & 0.000 & 0.000 & 0.000 & 0.000 \\
 & $M_1{+}C^{(300)}$ & 0.005 & 0.093 & 0.95 / 0.01 & 0.399 & 0.551 & $+0.393$ & $+0.458$ \\
 & $M_2$             & 0.000 & 0.000 & 1.00 / 1.00 & 0.000 & 0.000 & 0.000 & 0.000 \\
 & $M_2{+}C^{(100)}$     & 0.046 & 0.073 & 1.00 / 1.00 & 0.311 & 0.322 & $+0.266$ & $+0.249$ \\
 & $M_2{+}C^{(300)}$     & 0.192 & 0.533 & 1.00 / 1.00 & 0.868 & 0.971 & $+0.676$ & $+0.438$ \\
\hline
\end{tabular*}
\caption{Pre- and post-RL performance on 150 held-out compositions (64 samples per problem). $F$ / $G$ is pre-RL execution accuracy on isolated single-operation prompts with held-out strings. Post-RL values and changes are averaged over three GRPO seeds.}
\label{tab:empirical}
\end{table*}

\paragraph{Results.}
All three $M_2$ checkpoints execute the individual primitives with near-perfect accuracy yet have no observed successful trajectories on the held-out compositions before RL. Qwen3-1.7B and Qwen2.5-1.5B do not improve after RL with GRPO, while Gemma does. A post-hoc examination of the Gemma training logs showed that it had rare successful trajectories on the RL training prompts (6 out of 2,000 prompts solved at least once in 64 pre-RL samples), despite zero observed held-out pass@64. This highlights that effective support depends on the training distribution and sampling policy, rather than held-out benchmark performance alone.

The $M_1$ checkpoints, trained only on $F$, have zero success before and after RL across all three architectures. Composition supervision changes this outcome, but its effect depends on the amount of data provided. With 100 composition examples, $M_1{+}C^{(100)}$ still shows no improvement. With 300 examples, the checkpoints acquire nonzero full-solution support, and GRPO raises pass@1 to 0.17-0.40. Interestingly, these checkpoints retain near-zero accuracy on isolated $G$ probes, despite successfully using $G$ within some complete trajectories.

More generally, composition-focused supervision increases effective full-solution support and enables larger subsequent RL gains. When initial support is limited, RL can increase both pass@1 and pass@64, making previously rarely sampled solutions more accessible at finite sampling budgets. At high initial support, as with $M_2{+}C^{(300)}$ for Qwen3-1.7B, improvements are concentrated more heavily at pass@1 which is consistent with amplification.

\section{Limitations and Future Work}

\paragraph{Primitive coverage and composition.}
Our experiments show that isolated primitive mastery does not necessarily translate into successful composition. Models may execute individual primitives reliably while struggling to select and transition between them, and isolated probes may fail to reveal skills available in compositional contexts. Consequently, we do not yet establish that measured primitive coverage adds predictive power beyond full-solution support. Moreover, coverage estimates depend on how solutions are decomposed into primitives and the scaffolding used to probe them. A promising direction is to better characterise \emph{compositional connectivity} which can be thought of as the model's ability to select, sequence, and transition between available skills. Such orchestration capabilities may help explain variation within the composition-limited regime.

\paragraph{Support measurement and threshold selection.}
Effective full-solution support depends on the task distribution, sampling policy, and exploration budget. Failure to observe a successful trajectory at finite $k$ does not establish that it is absent, and our experiments show that held-out pass@$k$ can underestimate the reward accessibility encountered during RL training. Our classification also depends on thresholds $\tau_p$ and $\tau_c$, which we do not yet offer a principled selection procedure for. Future work could calibrate these thresholds to the training-time exploration budget or replace discrete regimes with continuous models of post-training outcomes.

\paragraph{Generalisability.}
Our experiments use small language models and synthetic composition tasks with simple and known primitives and deterministic verification. While this allows us to isolate different capability bottlenecks in a toy example, natural reasoning tasks admit multiple valid decompositions, less clearly defined primitives, and more complex forms of composition. Extending the framework to mathematical reasoning, coding, and larger models will be important for establishing whether these diagnostics generalise beyond controlled settings.

\section{Conclusion}
We have argued that the question ``does RL create new capabilities?'' is ill-posed when asked in the abstract. It is better replaced by a conditional question: given a base model’s capability state on a task, what can post-training do from there? The state space we propose factors the state into two measurable properties -- effective full-solution support and atomic primitive coverage -- that aggregate model accuracy conflates. Apparent contradictions in the RLVR literature can then be read as observations about different regions of this space: pass@k convergence results characterise amplification, controlled composition studies characterise the composition-limited regime, and failures of vanilla RL on hard tasks may reflect acquisition-limited states or composition-limited states in which complete solutions are too rare to be rewarded. Our synthetic experiments support the central role of effective support. Outcome-based RL improved only checkpoints with some reachable successful trajectories and composition supervision that created such trajectories subsequently enabled large gains. They also show that primitive mastery is not sufficient on its own and that measured support depends on the distribution and budget under which it is probed. The two axes are not independent because any complete solution must be built from a valid sequence of operations, and the space carries internal structure that can be used to validate measurements rather than merely classify outcomes.

\clearpage
\bibliographystyle{abbrvnat}
\bibliography{references}

\end{document}